\documentclass[letterpaper]{article} % DO NOT CHANGE THIS
\usepackage{aaai23}  % DO NOT CHANGE THIS
\usepackage{times}  % DO NOT CHANGE THIS
\usepackage{helvet}  % DO NOT CHANGE THIS
\usepackage{courier}  % DO NOT CHANGE THIS
\usepackage[hyphens]{url}  % DO NOT CHANGE THIS
\usepackage{graphicx} % DO NOT CHANGE THIS
\usepackage{natbib}  % DO NOT CHANGE THIS AND DO NOT ADD ANY OPTIONS TO IT
\usepackage{caption} % DO NOT CHANGE THIS AND DO NOT ADD ANY OPTIONS TO IT
\usepackage{algorithm}
\usepackage{algorithmic}
\usepackage{amsmath}
\usepackage{amssymb}
\usepackage{amsfonts}
\usepackage{mathtools}
\usepackage{newfloat}
\usepackage{listings}
\DeclareCaptionStyle{ruled}{labelfont=normalfont,labelsep=colon,strut=off} % DO NOT CHANGE THIS
\floatstyle{ruled}
\newfloat{listing}{tb}{lst}{}
\floatname{listing}{Listing}
\usepackage{pifont}% http://ctan.org/pkg/pifont
\newcommand{\cmark}{\ding{51}}%
\newcommand{\xmark}{\ding{55}}%

\title{Fourier-Net: Fast Image Registration with Band-Limited Deformation}
\author {
    Xi Jia\textsuperscript{\rm 1},
    Joseph Bartlett\textsuperscript{\rm 1,2},
    Wei Chen\textsuperscript{\rm 1},
    Siyang Song\textsuperscript{\rm 3}, 
    Tianyang Zhang\textsuperscript{\rm 1}, \\
    Xinxing Cheng\textsuperscript{\rm 1}, 
    Wenqi Lu\textsuperscript{\rm 4}, 
    Zhaowen Qiu\textsuperscript{\rm 5{$\dagger$}}, 
    Jinming Duan\textsuperscript{\rm 1,6{$\dagger$}}
}
\affiliations {
    \textsuperscript{\rm 1} School of Computer Science, University of Birmingham, UK\\
    \textsuperscript{\rm 2} Department of Biomedical Engineering, University of Melbourne, Australia \\
    \textsuperscript{\rm 3}  Department of Computer Science and Technology, University of Cambridge, UK \\
    \textsuperscript{\rm 4} Department of Computer Science, University of Warwick, UK \\
    \textsuperscript{\rm 5} Institute of Information Computer Engineering, Northeast Forestry University, China \\
    \textsuperscript{\rm 6} Alan Turing Institute, UK \\
}
\begin{document}

\maketitle

\begin{abstract}
Unsupervised image registration commonly adopts U-Net style networks to predict dense displacement fields in the full-resolution spatial domain. For high-resolution volumetric image data, this process is however resource-intensive and time-consuming. To tackle this problem, we propose the Fourier-Net, replacing the expansive path in a U-Net style network with a parameter-free model-driven decoder. Specifically, instead of our Fourier-Net learning to output a full-resolution displacement field in the spatial domain, we learn its low-dimensional representation in a band-limited Fourier domain. This representation is then decoded by our devised model-driven decoder (consisting of a zero padding layer and an inverse discrete Fourier transform layer) to the dense, full-resolution displacement field in the spatial domain. These changes allow our unsupervised Fourier-Net to contain fewer parameters and computational operations, resulting in faster inference speeds. Fourier-Net is then evaluated on two public 3D brain datasets against various state-of-the-art approaches. For example, when compared to a recent transformer-based method, named TransMorph, our Fourier-Net, which only uses 2.2\% of its parameters and 6.66\% of the multiply-add operations, achieves a 0.5\% higher Dice score and an 11.48 times faster inference speed. Code is available at \url{https://github.com/xi-jia/Fourier-Net}.

%  that learns a band-limited representation of the deformation to speed up the inference while maintaining the registration performance, which we term

%our Fourier-Net only requires 18\% multiply-adds operations and achieve 2.7x inference speed with 3.3\% DiceDice score improvement over the state-of-the-art

% We evaluate our Fourier-Net on a 2D and a 3D cardiac datasets, the results achieved on the 3D 

% on the 3D setting, Fourier-Net outperforms the state of the art by 3.3\% in terms of DiceDice score with only 18\% multiply-adds operations, resulting in 2.7x inference speed.

 %the results show that Fourier-Net is faster and is able to achieve a competitive performance with state-of-the-art methods, especailly
 
% \keywords{Image registration \and Fourier transformation \and Low dimensional deformation \and Band-limited deformation.}

\end{abstract}

\section{Introduction} 
Medical image registration aims to learn a spatial deformation that identifies the correspondence between a moving image and a fixed image, which is a fundamental step in many medical image analysis applications such as longitudinal studies, population modeling, and statistical atlases \cite{sotiras2013deformable}.

Iterative optimization techniques such as FFD \cite{rueckert1999nonrigid}, Demons \cite{vercauteren2009diffeomorphic}, ANTS \cite{avants_ANTS}, Flash \cite{zhang2019fast} and ADMM \cite{thorley2021nesterov} have been applied to deformable image registration. However, such optimization-based approaches require elaborate hyper-parameter tuning for each image pair, and iteration towards an optimal deformation is very time-consuming, thus limiting their applications in real-time and large-scale volumetric registration.

\begin{figure}[t]
    \centering
    \includegraphics[width=0.45\textwidth]{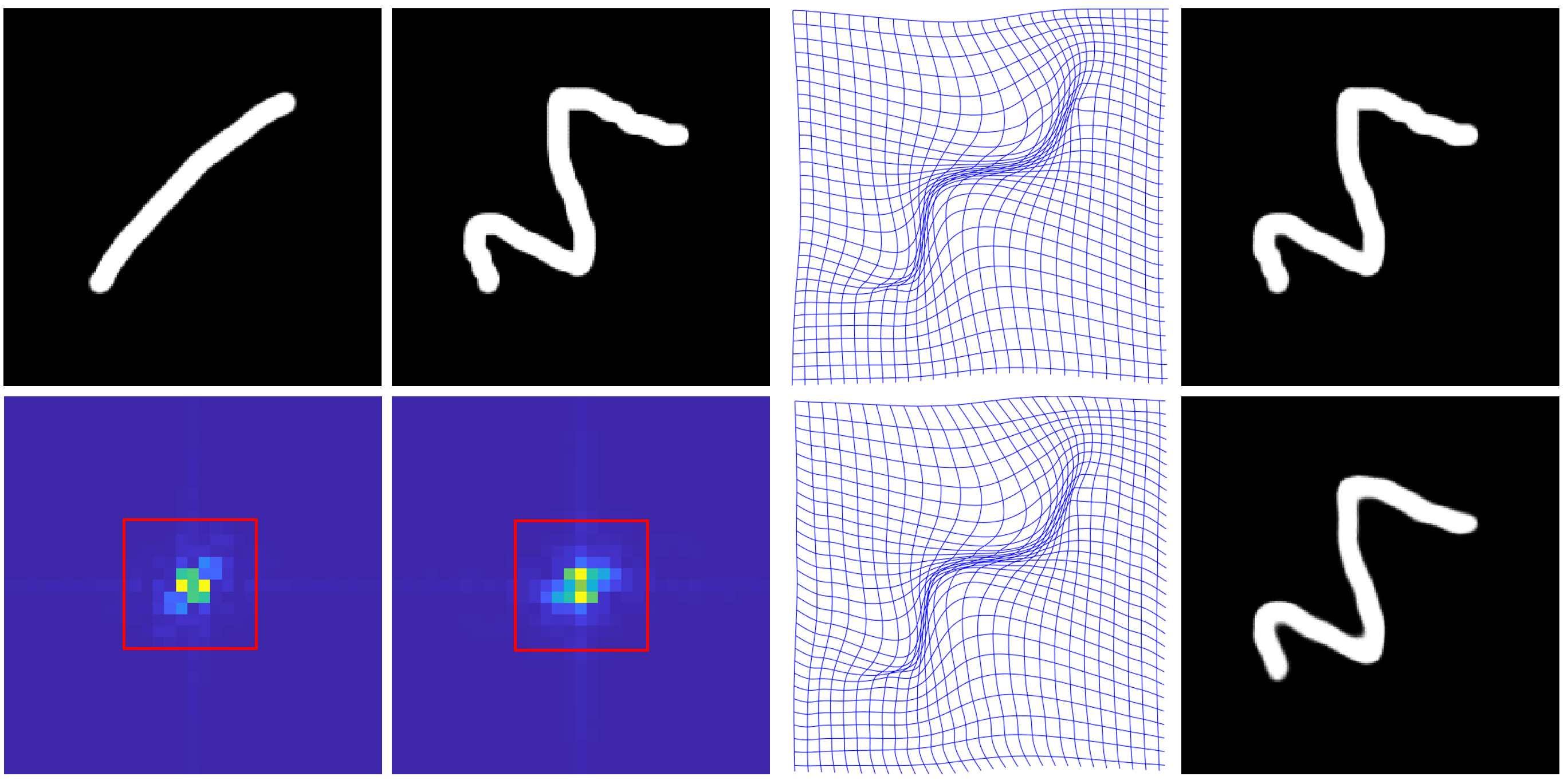}
    \caption{From left to right in the 1st row: moving image, fixed image, deformation grid, and warped moving image. From left to right in the 2nd row: DFT of horizontal displacement field, DFT of vertical displacement field, reconstructed deformation grid from Fourier coefficients only in band-limited region (marked by red rectangles), and the warped moving image by reconstructed deformation.}
    \label{fig:one}
\end{figure}

In recent years, deep learning based approaches have burgeoned in the field of medical image registration \cite{hering2022learn2reg}. Their success has been largely driven by their exceptionally fast inference speeds. The most effective methods, such as VoxelMorph \cite{balakrishnan2019voxelmorph}, usually adopt a U-Net style architecture to estimate dense, spatial deformation fields. They only require one forward propagation during inference, and thus can register images several orders of magnitudes faster than traditional iterative methods. Following the success of VoxelMorph, a large number of deep learning based approaches have been proposed for various registration tasks \cite{zhang2018inverse, Zhao_2019_ICCV, Mok_2020_CVPR, jia2021learning, kim2021cyclemorph, chen2021transmorph, jia2022u}. These models either use multiple U-Net style networks in a cascaded way or replace basic convolution blocks in VoxelMorph with more sophisticated ones such as swin-transformers \cite{chen2021transmorph} to boost registration performance. However, these changes rapidly increase the number of network parameters and multiply-add operations (mult-adds), sacrificing training and inference efficiency altogether.

For U-Net-based registration models, we argue 1) that it may be unnecessary to include the whole expansive path of U-Net backbone and 2) that the training and inference efficiency of such networks can be further improved by learning a low-dimensional representation of displacement field in a band-limited Fourier domain. Our arguments are based on our observation in Figure~\ref{fig:one}, where we notice it is sufficient to reconstruct an accurate full-resolution deformation (the third image of the second row in Figure~\ref{fig:one}) by using only a small number of coefficients in the band-limited Fourier domain. Inspired by this insight, we propose an end-to-end unsupervised approach that learns only a low-dimensional representation of displacement field in the band-limited Fourier domain. We term our approach the Fourier-Net.

By removing several layers in the expansive path of a U-Net style architecture, our Fourier-Net outputs only a small patch that stores low-frequency coefficients of displacement field in the Fourier domain. We then propose to directly apply a model-driven decoder to recover the full-resolution spatial displacement field from these low-frequency Fourier coefficients. This model-driven decoder contains a zero-padding layer that broadcasts complex-valued low-frequency signals into a full-resolution complex-valued map. The inverse discrete Fourier transform (iDFT) is then applied to this map to obtain the full-resolution spatial displacement field. Both zero-padding and iDFT layers are parameter-free and therefore fast. On top of Fourier-Net, we propose a diffeomorphic variant, termed Fourier-Net-Diff. This network first estimates a stationary velocity field, followed by squaring and scaling layers \cite{dalca2018unsupervised} to encourage the output deformation to be diffeomorphic.

\section{Related Works}

\textbf{Unsupervised approaches} can be based on either iterative optimization or learning. Iterative methods are prohibitively slow, especially when the images to be registered are of a high-dimensional form, such as 3D volumes. Over the past decades, many works have been proposed to accelerate such methods. \cite{ashburner2007fast} used a stationary velocity field (SVF) representation \cite{legouhy2019unbiased}, and proposed a fast algorithm DARTEL for image registration which computes the resulting deformation by using scaling and squaring from the SVF. Another fast approach for image registration is Demons \cite{vercauteren2009diffeomorphic}, which imposes smoothness on displacement fields by incorporating inexpensive Gaussian convolutions into its iterative process. Hernandez \cite{hernandez2018band} reformulated the Stokes-LDDMM variational problem used in \cite{mang2015inexact} in the domain of band-limited non-stationary vector fields and utilized GPUs to parallelize their methods. \cite{zhang2019fast} developed the Fourier-approximated Lie algebras for shooting (Flash) for fast diffeomorphic image registration, where they proposed to speed up the solution of the Euler-Poincaré differential (EPDiff) equation used to compute deformations from velocity fields in the band-limited Fourier domain.

On the other hand, deep learning methods based on convolutional neural networks have been employed to overcome slow registration speeds. Among them, U-Net style networks have been proven to be an effective tool to learn deformations between pairwise images \cite{balakrishnan2019voxelmorph,zhang2018inverse,Mok_2020_CVPR,kim2021cyclemorph}.  While their registration performance is comparable with iterative methods, their inference can be orders of magnitude faster. RC-Net \cite{Zhao_2019_ICCV} and VR-Net \cite{jia2021learning} cascaded multiple U-Net style networks to improve the registration performance, but their speed is relatively slow. Very recently, approaches, such as  ViT-V-
Net \cite{chen2021vit} and TransMorph \cite{chen2021transmorph}, which combine vision transformers and U-Nets have achieved promising registration performance, but they involve much more computational operations and are therefore slow. Another group of network-based image registration techniques \cite{de2019deep,qiu2021learning} is to estimate a grid of B-Spline control points with regular spacing, which is then interpolated based on cubic B-Spline basis functions \cite{rueckert1999nonrigid, duan2019automatic}. By estimating fewer control points, these networks perform fast predictions, but currently are less accurate.

\textbf{Supervised approaches} are also studied in medical image registration. However, they have several pitfalls: 1) it is generally hard to provide human-annotated ground truth deformations for supervision; and 2) if trained using numerical solutions of other iterative methods, the performance of these supervised registration methods may be limited by iterative methods. Yang et al. proposed Quicksilver \cite{yang2017quicksilver} which is a supervised encoder-decoder network and trained using the initial momentum of LDDMM as the supervision signal. Wang et al. extended Flash \cite{zhang2019fast} to DeepFlash \cite{wang2020deepflash} in a learning framework in lieu of iterative optimization. Compared to Flash, DeepFlash accelerates the computation of initial velocity fields but needs to solve a PDE (i.e., EPDiff equation) in the Fourier domain so as to recover the full-resolution deformation in the spatial domain, which can be slow. The fact that DeepFlash requires the numerical solutions of Flash \cite{zhang2019fast} as training data attributes to lower registration performance than Flash.

 % Due to its fast patch-wise training strategy, Quicksilver can be trained swiftly and efficiently.

\begin{figure*}[t]
     \centering
     \includegraphics[width=0.96\textwidth]{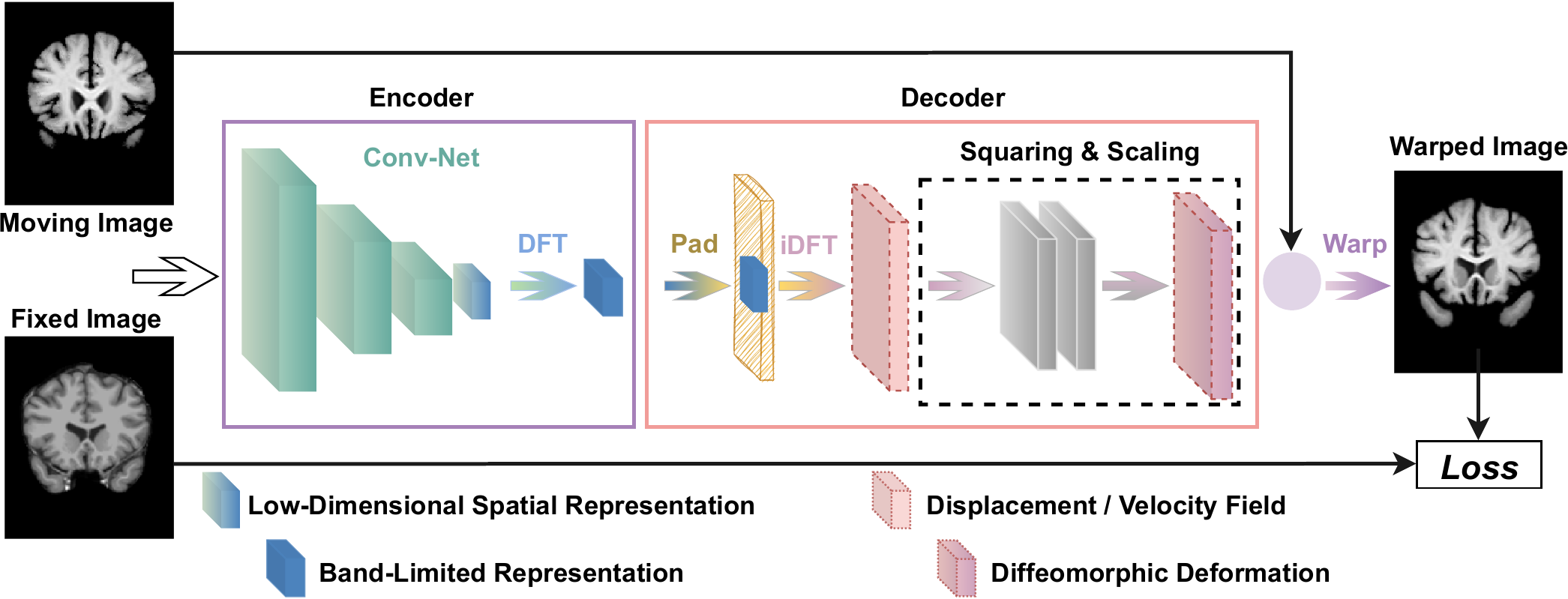}
     \caption{Architecture of our end-to-end Fourier-Net. It contains 1) a convolutional encoder that first produces a low-dimensional representation of displacement or velocity field, followed by an embedded discrete Fourier transformation (DFT) layer to map this low-dimensional representation into the band-limited Fourier domain; 2) a parameter-free model-driven decoder that adopts a zero-padding layer, an inverse DFT (iDFT) layer, and seven optional squaring and scaling layers to reconstruct the displacement field or deformation into the full-resolution spatial domain from its band-limited Fourier domain; 3) a warping layer to deform the moving image; and 4) a loss function that includes a similarity term and a regularization term.  }
     \label{fig:flowchart}
 \end{figure*}

Although DeepFlash also learns a low-dimensional band-limited representation, it differs from our Fourier-Net in four aspects, which we reckon our novel contributions to this area. First, DeepFlash is a supervised method that requires ground truth velocity fields calculated from Flash prior to training, whilst Fourier-Net is a simple and effective unsupervised method thanks to our proposed model-driven decoder. Second, DeepFlash is a multi-step method whose network's output requires an additional PDE algorithm \cite{zhang2019fast} to compute final full-resolution spatial deformations, whilst Fourier-Net is a holistic model that can be trained and used in an end-to-end manner. Third, DeepFlash needs two individual convolutional networks to estimate real and imaginary signals in the band-limited Fourier domain, whilst Fourier-Net uses only one single network directly mapping image pairs to a reduced-resolution displacement field without the need of complex-valued operations. Lastly, DeepFlash is essentially an extension of Flash and it is difficult for the method to benefit from vast amounts of data, whilst Fourier-Net is flexible and can easily learn from large-scale datasets. Due to these, our Fourier-Net outperforms DeepFlash (as well as Flash) by a significant margin in terms of both accuracy and speed.

\section{Methodology}
% In this section, we detail the proposed end-to-end unsupervised Fourier-Net.
As illustrated in Figure~\ref{fig:flowchart}, in Fourier-Net, the encoder first takes a pair of spatial images as input, and encodes them to a low-dimensional representation of displacement field (or velocity field if diffeomorphisms are imposed) in the band-limited Fourier domain. Then the decoder brings the displacement field (or velocity field) from the band-limited Fourier domain to the spatial domain, and ensures that they have the same spatial size as the input image pair. Next, the optional squaring and scaling layers are used to encourage a diffeomorphism in final deformations. Finally, by minimizing the loss function, an accurate deformation can be estimated, with which the warping layer deforms the moving image to be similar to the fixed image.  

\subsection{Encoder}
The encoder aims to learn a displacement or velocity field in the band-limited Fourier domain. Intuitively, this may require convolutions to be able to handle complex-valued numbers. One may directly use complex convolutional networks \cite{trabelsi2017deep}, they are suitable when both input and output are complex values, but complex-valued operations sacrifice computational efficiency. Instead, DeepFlash \cite{wang2020deepflash} tackles this problem by first converting input image pairs to the Fourier domain and then using two individual real-valued convolutional networks to learn the real and imaginary signals separately. Such an approach increase the training and inference cost (as listed in Table \ref{tab:ablation}).  Since our Fourier-Net estimates displacement fields in the band-limited Fourier domain from spatial images (inputs are real values but outputs are complex values), these approaches may not be well suited to our application.
 
\begin{figure}[ht]
    \centering
    \includegraphics[width=0.4\textwidth]{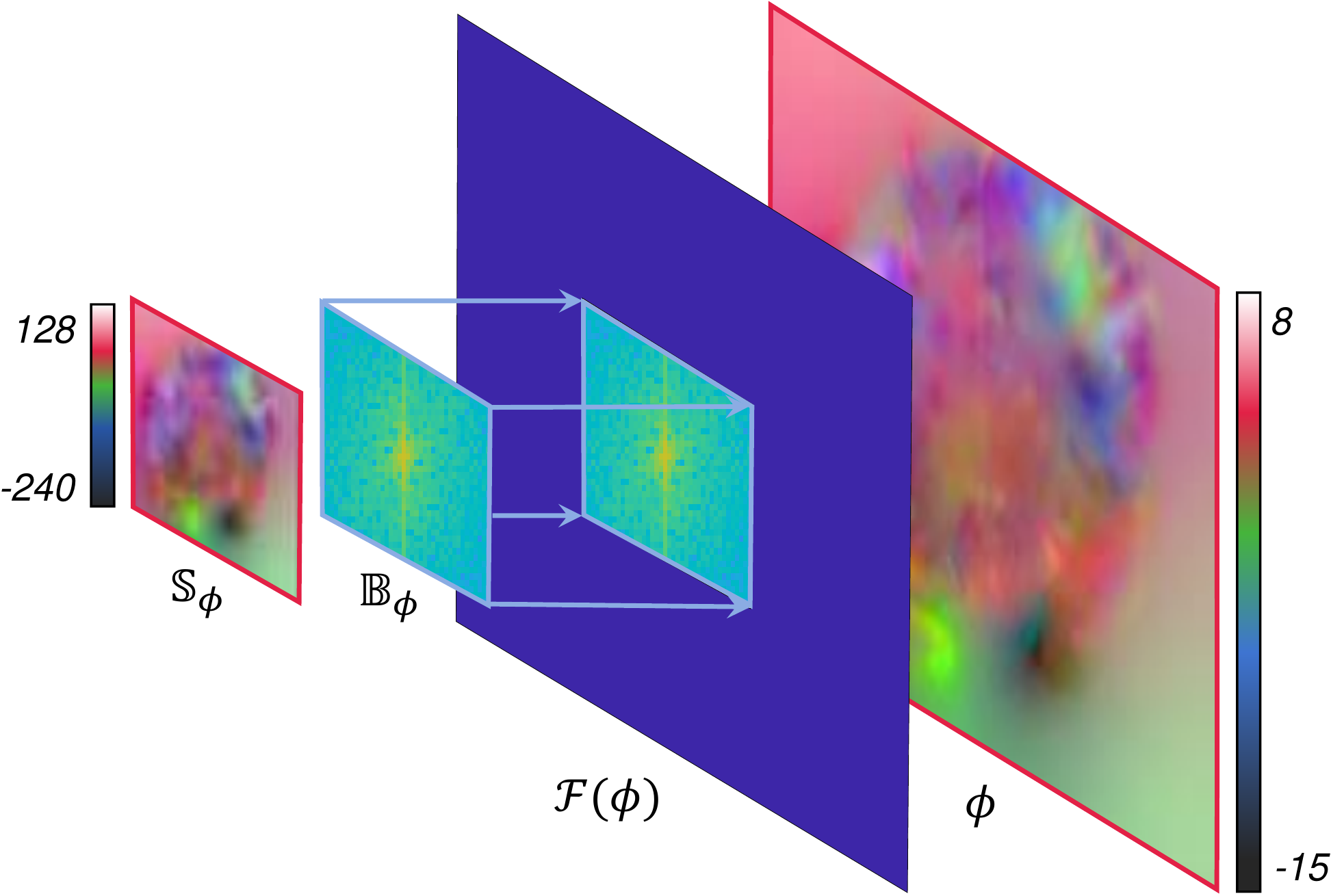}
    \caption{Connection between low-dimensional spatial displacement field $\mathbb{S}_{\boldsymbol{\phi}}$, band-limited Fourier coefficients $\mathbb{B}_{\boldsymbol{\phi}}$, full-resolution Fourier coefficients $\cal{F}({\boldsymbol{\phi}})$ by zero-padding $\mathbb{B}_{\boldsymbol{\phi}}$, and full-resolution displacement field $\boldsymbol{\phi}$ by taking iDFT of $\cal{F}(\boldsymbol{\phi})$.} \label{fig:large_small_spatial}
\end{figure}

To bridge the domain gap between real-valued spatial images and complex-valued band-limited displacement fields without increasing complexity, we propose to embed a DFT layer at the end of the convolutional network in the encoder. This is a simple and effective way to produce complex-valued band-limited displacement fields without the network being able to handle complex values itself. Let us denote the moving image as $I_0$, the fixed image as $I_1$, the convolutional network as CNN with the parameters $\bf{\Theta}$, the DFT as $\cal F$, the full-resolution spatial displacement field as $\boldsymbol{\phi}$, and the complex band-limited displacement field as $\mathbb{B}_{\boldsymbol{\phi}}$. In this case, our encoder can be defined as $\mathbb{B}_{\boldsymbol{\phi}} = {\cal F}({\rm{CNN}}(I_0, I_1; {\bf{\Theta}}))$, resulting in a compact, efficient implementation as compared to DeepFlash and other complex convolutional networks. On the other hand, we also notice from our experiments (Table \ref{tab:ablation}) that it is difficult to regress $\mathbb{B}_{\boldsymbol{\phi}}$ directly from $I_0$ and $I_1$ within a single CNN, i.e., $\mathbb{B}_{\boldsymbol{\phi}} = {\rm{CNN}}(I_0, I_1; {\bf{\Theta}})$. We believe the reason being that: if the CNN directly learns a band-limited displacement field, it needs to go through two domains altogether: first mapping the spatial images to the spatial displacement field and then mapping this displacement field into its band-limited Fourier domain. In this case, the domain gap is too big. Our network however only needs to go through one domain and then DFT handles the second domain. By doing so, Fourier-Net is efficient and easy to learn. An illustration of this idea is given in Figure~\ref{fig:large_small_spatial}.

So far, we have given an intuitive explanation of how the encoder in our network learns. Here we discuss their mathematical relationship between the low-dimensional spatial displacement field $\mathbb{S}_{\boldsymbol{\phi}}= {\rm{CNN}} (I_0, I_1; {\bf{\Theta}})$, its band-limited representation $\mathbb{B}_{\boldsymbol{\phi}}$, as well as the displacement field $\boldsymbol{\phi}$ (coming after the decoder) in the full-resolution spatial domain. For simplicity, we use a 2D displacement field as an example and the formulations below can be easily extended to 3D cases. A general discrete Fourier transform used on $\boldsymbol{\phi}$ can be defined as follows:

\begin{equation} \label{eq:dft}
[{\cal {F}}(\boldsymbol{\phi})]_{k, l}= \sum_{i=0}^{M-1} \sum_{j=0}^{N-1} \boldsymbol{\phi}_{i, j} e^{-\sqrt{-1} \left(\frac{2 \pi k}{M} i+\frac{2 \pi l}{N} j\right)},
\end{equation}
where $\boldsymbol{\phi}$ is of size $M\times N$, $i \in [0, M-1]$ and $j\in [0, N-1]$  are the discrete indices in the spatial domain, and $k \in [0, M-1]$ and $l \in [0, N-1]$ are the discrete indices in the frequency domain.

In our Fourier-Net, $\boldsymbol \phi$ is actually a low-pass filtered displacement field. If we define a $M \times N$ sized sampling mask $\cal D$ whose entries are zeros if they are on the positions of high-frequency signals in $\boldsymbol \phi$ and ones if they are on the low-frequency positions. With $\cal D$, we can recover the displacement field $\boldsymbol{\phi}$ from Eq.~\eqref{eq:dft}
\begin{equation} \label{eq:idft}
 {\boldsymbol{\phi}}_{i, j} = {\frac{1}{MN}}\sum_{k=0}^{M-1} \sum_{l=0}^{N-1} {\cal{D}}_{k,l} [{\cal {F}}(\boldsymbol{\phi})]_{k, l} e^{\sqrt{-1} \left(\frac{2 \pi i}{M} k+\frac{2 \pi j}{N} l\right)}.
\end{equation}

If we shift all low-frequency signals of the displacement field to a center patch of size $\frac{M}{a}\times\frac{N}{b}$ ($\frac{M}{a},\frac{N}{b},a=2Z_a,b=2Z_b,Z_a,Z_b \in \mathbb{Z}^+$), center-crop the patch (denoted by $\mathbb{B}_{\boldsymbol{\phi}}$), and then perform the iDFT on this patch, we obtain $\mathbb{S}_{\boldsymbol{\phi}}$ in Eq.~\eqref{eq:dft2}

\begin{equation} \label{eq:dft2}
[\mathbb{S}_{\boldsymbol{\phi}}]_{\widehat{i}, \widehat{j}} = \frac{ab}{M N}  \sum_{\widehat{k}=0}^{\frac{M}{a}-1} \sum_{\widehat{l}=0}^{\frac{N}{b}-1} [\mathbb{B}_{\boldsymbol{\phi}}]_{\widehat{k},\widehat{l}} e^{\sqrt{-1} \left(\frac{2 \pi a\widehat{i}}{M} \widehat{k}+\frac{2 \pi b\widehat{j}}{N} \widehat{l}\right)},
\end{equation}
where $\widehat{i} \in [0, \frac{M}{a}-1]$ and $\widehat{j}\in [0, \frac{N}{b}-1]$ are the indices in the spatial domain, and $\widehat{k} \in [0, \frac{M}{a}-1]$ and $\widehat{l} \in [0, \frac{N}{b}-1]$ are the indices in the frequency domain. Note that $\mathbb{S}_{\boldsymbol{\phi}}$ is a low-dimensional spatial representation of $\boldsymbol \phi$ and we are interested in their mathematical connection. Another note is that $\mathbb{S}_{\boldsymbol{\phi}}$ actually contains all the information of its band-limited Fourier coefficients in $\mathbb{B}_{\boldsymbol{\phi}}$. As such, we do not need the network to learn the coefficients in $\mathbb{B}_{\boldsymbol{\phi}}$ and instead only to learn its real-valued coefficients in $\mathbb{S}_{\boldsymbol{\phi}}$.

Since most of entries ($\frac{a \times b -1}{a \times b}\%$) in $\cal {F}(\boldsymbol{\phi})$ are zeros, and the values of rest entries are exactly the same as in $\mathbb{B}_{\boldsymbol{\phi}}$, we can conclude that $\mathbb{S}_{\boldsymbol{\phi}}$ contains all the information $\boldsymbol{\phi}$ can provide, and their mathematical connection is
\begin{equation}
[\mathbb{S}_{\boldsymbol{\phi}}]_{\widehat{i}, \widehat{j}}  =  ab \times \boldsymbol{\phi}_{a \widehat{i}, b\widehat{j}}.
\end{equation}
With this derivation, we show that we can actually recover a low-dimensional spatial representation $\mathbb{S}_{\boldsymbol{\phi}}$ from its full-resolution spatial displacement field $\boldsymbol{\phi}$, as long as they have the same low-frequency coefficients $\mathbb{B}_{\boldsymbol{\phi}}$. This essentially proves that there exists a unique mapping function between $\mathbb{S}_{\boldsymbol{\phi}}$ and $\boldsymbol{\phi}$ and that it is reasonable to use a network to learn $\mathbb{S}_{\boldsymbol{\phi}}$ directly from image pairs.

\subsection{Model-Driven Decoder}
The proposed decoder consists of a zero-padding layer, an iDFT layer, and an optional squaring and scaling module.
%We detail each in the following.

The output from the encoder is a band-limited representation $\mathbb{B}_{\boldsymbol{\phi}}$. To recover the full-resolution displacement field $\boldsymbol{\phi}$ in the spatial domain, we first pad the patch $\mathbb{B}_{\boldsymbol{\phi}}$ containing mostly low-frequency signals to the original image resolution with zero values (i.e., ${\cal {F}}(\boldsymbol{\phi})$). We then feed the zero-padded complex-valued coefficients ${\cal {F}}(\boldsymbol{\phi})$ to an iDFT layer consisting of two steps: shifting the Fourier coefficients from centers to corners and then applying the standard iDFT to convert them into the spatial domain. The output from Fourier-Net is thus a full-resolution spatial displacement field. Both padding and iDFT layers are differentiable and therefore Fourier-Net can be optimized via standard back-propagation. We note that our proposed decoder is a parameter-free module that is driven by knowledge instead of learning and therefore fast.

% In addition to computational advantages, the use of band-limited parameterization in the Fourier domain acts as a low-pass filter, thus inducing intrinsic smoothness to resulting deformations.
% This small patch can be treated as the low-frequency components of a smooth deformation in the band-limited Fourier domain.

We also propose a diffeomorphic variant of Fourier-Net which we term Fourier-Net-Diff. A diffeomorphic deformation is defined as a smooth and invertible deformation, and in Fourier-Net-Diff we need an extra squaring and squaring module for the purpose. The output of the iDFT layer can be regarded as a stationary velocity field denoted by $\boldsymbol{v}$ instead of the displacement field $\boldsymbol \phi$. In group theory, $\boldsymbol{v}$ is a member of Lie algebra, and we can exponentiate this stationary velocity field (i.e., $Exp(\boldsymbol{v}$)) to obtain a diffeomorphic deformation. In this paper, we use seven scaling and squaring layers \cite{ashburner2007fast,dalca2018unsupervised} to impose such a diffeomorphism.

\subsection{Warping Layer and Loss Functions}

After the {model-driven decoder}, we obtain a full-resolution displacement field (or a diffeomorphic deformation) for the input image pair. We then deform the moving image using a warping layer to produce the warped moving image, which is then used to calculate the loss. We implement 2D and 3D spatial warping layers based on linear interpolation as in \cite{jaderberg2015spatial} and \cite{balakrishnan2019voxelmorph}.

We adopt an unsupervised loss which is computed from the moving image $I_1$, the fixed image $I_0$, and the predicted displacement field $\boldsymbol{\phi}$ or velocity field $\boldsymbol{v}$. The training objective of our Fourier-Net is ${\cal L}({\bf{\Theta}}) = \min_{\bf{\Theta}}  \frac{1}{N} \sum_{i=1}^N {\cal L}_{Sim}( I_1^i \circ  ({\boldsymbol{\phi}_i}({\bf{\Theta}}) + {\rm{Id}}) - I_0^i )  + \frac{\lambda}{N} \sum_{i=1}^N \|\nabla \boldsymbol{\phi}_i({\bf{\Theta}}) \|_2^2 $,
where $N$ is the number of training pairs, ${\bf{\Theta}}$ is the network parameters to be learned, ${\rm{Id}}$ is the identity grid, $\circ$ is the warping operator, and $\nabla$ is the first order gradient implemented using finite differences \cite{lu2016implementation,duan2016edge}. The first term ${\cal L}_{Sim}$ defines the similarity between warped moving images and fixed images, and the second term defines the smoothness of displacement fields. Here $\lambda$ is a hyper-parameter balancing the two terms. As for Fourier-Net-Diff, the training loss is defined as ${\cal L}({\bf{\Theta}})=\min_{\bf{\Theta}}  \frac{1}{N} \sum_{i=1}^N  {\cal L}_{Sim}( I_1^i \circ  Exp(\boldsymbol{v}_i({\bf{\Theta}})) - I_0^i )  + \frac{\lambda}{N} \sum_{i=1}^N \|\nabla \boldsymbol{v}_i({\bf{\Theta}}) \|_2^2 $. ${\cal L}_{Sim}$ can be either mean squared error (MSE) or normalized cross-correlation (NCC), which we clarify in our experiments.

\section{Experiments}
\subsection{Datasets}
% We use two publicly available brain datasets to evaluate our proposed Fourier-Net and Fourier-Net-Diff.
\textbf{OASIS-1 dataset} \cite{marcus2007open} consists of a cross-sectional collection of T1-weighted brain MRI scans from 416 subjects. In experiments, we use the pre-processed OASIS data\footnote{\url{https://learn2reg.grand-challenge.org/Learn2Reg2021/}} provided by \cite{hoopes2021hypermorph} to perform subject-to-subject brain registration, in which all 414 MRI scans are bias-corrected, skull-stripped, aligned, and cropped to the size of $160\times192\times224$. Automated segmentation masks from FreeSurfer are provided for evaluation of registration accuracy. This dataset also has 414 2D slices and marks extracted from their corresponding 3D volumes. We randomly split this 2D dataset into 201, 12, and 201 images for training, validation, and test. After pairing, we end up with 40200 (201$\times$200), 22 ([12-1]$\times$2), and 400 ([201-1]$\times$2) image pairs for training, validation, and test, respectively.
 % the MICCAI 2021 Learn2Reg challenge
%We use function \textit{product} and \textit{zip} in \textit{itertools} to implement the fast pair construction.
 % which is also used in the MICCAI 2021 Learn2Reg Challenge 

\textbf{IXI dataset}\footnote{\url{https://brain-development.org/ixi-dataset/}} contains nearly 600 MRI scans from healthy subjects. In experiments, we use the pre-processed IXI data provided by \cite{chen2021transmorph} to perform atlas-based brain registration. The atlas is generated by \cite{kim2021cyclemorph}. There are in total 576 $160\times192\times224$ 3D brain MRI volumes in this dataset. The dataset is split into 403 for training, 58 for validation, and 115 for testing. There is no pairing step for this dataset as it is an atlas-to-subject registration task.

\subsection{Implementation Details}
We implement our Fourier-Net using PyTorch, where training is optimized using Adam with a fixed learning rate of 0.0001. We tune built-in hyper-parameters on a held-out validation set. Specifically, we use MSE to train both 2D and 3D OASIS for 10 and 1000 epochs, respectively,  and $\lambda$ in ${\cal L}({\bf{\Theta}})$ is set to 0.01. For 3D OASIS, an additional Dice loss is used with its weight being set to 1. On 3D IXI, we train the models with NCC loss for 1000 epochs with $\lambda=$ 5. All deep models are trained with an Nvidia A100 GPU.

The CNN in Fourier-Net has 6 convolutional blocks. The initial 4 blocks contain 2 convolutional layers in each block. The first layer maintains the same spatial resolution as inputs, while the second layer performs a down-sampling with a stride of 2 and then doubles the number of feature channels. In the last 2 blocks, each contains a fractional convolutional layer and 2 convolutional layers. The fractional layer performs an up-sampling with a stride of 2, and the convolutional layers halve the number of feature channels. The kernel size in all convolutional layers is 3$\times$3$\times$3. Each convolution is followed by a PReLU activation except the last sub-layer, which does not have any activation layer and contains 2 or 3 kernels for 2D or 3D registration, respectively. The initial number of kernels in the first convolutional layer is set to $C$. For example, the spatial resolution of input images changes from 160$\times$192$\times$224$\times$2 to
80$\times$96$\times$112$\times$C
$\rightarrow$
40$\times$48$\times$56$\times$2C
$\rightarrow$
20$\times$24$\times$28$\times$4C
$\rightarrow$
10$\times$12$\times$14$\times$8C
$\rightarrow$
20$\times$24$\times$28$\times$4C
$\rightarrow$
40$\times$48$\times$56$\times$3 after each block. We experiment $C$=8, 16, and 48, which define small Fourier-Net$_{\textbf{S}}$, Fourier-Net, and large Fourier-Net$_{\textbf{L}}$, respectively. Though the output of our Fourier-Net is set to 40$\times$48$\times$56, the resolution is not constrained and one can customize the CNN architecture to produce a band-limited representation with any resolution. To adapt Fourier-Net onto 2D images, we directly change all 3D kernels to 2D.

\subsection{Ablation Studies and Parameter Tuning} 
\label{subsec:ablation}

%Discrete Fourier Transform (DFT) is used to find the frequency domain.the objective of registration method is to learn a smooth deformation field, which naturally can be compressed in the frequency domain.

% In our applications, the output deformation is derived from the intermediate layers of the expansive path.
The first question we ask is what  the most suitable resolution  (i.e., patch size) of a band-limited displacement field is?
% There is a trade off between the registration performance and speed, i.e., smaller resolution will result in high speed but lower registration accuracy.
A very small patch will rapidly decrease model parameters as well as training and inference time but may lead to lower performance. A very large patch could retain registration accuracy but may increase training and inference time, thus eliminating the advantages of our method.

In Table \ref{tab:ablation}, we use 2D OASIS images for ablation studies and investigate the impact of different patch sizes, i.e, 20$\times$24 and 40$\times$48, which are respectively $\frac{1}{64}$ and $\frac{1}{16}$ of the original image size. It can be seen that the 40$\times$48 patch improves Dice by 2\% over the 20$\times$24 patch, with only a slight difference in mult-adds operations. The Dice score of our Fourier-Net (40$\times$48) is already close to the full-resolution U-Net backbone (last row in this Table), which however has 2.5 times mult-adds cost than our Fourier-Net (40$\times$48).

\begin{table}[t]
\centering
% \resizebox{\columnwidth}{!}{
\begin{small}
\begin{tabular}{cccccc}
\hline
Patch        & DFT        & SS & Dice$\uparrow$        & $|J|_{< 0}\%$      & MA(M) \\\hline
20$\times$24      & \xmark  & \xmark       & .664$\pm$.040  & .158$\pm$.206 & 891\\
20$\times$24      & \cmark  & \xmark       & .732$\pm$.042  & .434$\pm$.355 & 679 \\
20$\times$24      & \cmark  & \cmark       & .735$\pm$.037  & 0.0$\pm$0.0     & 679 \\
40$\times$48      & \xmark  & \xmark       & .675$\pm$.038  & .279$\pm$.257 & 1310\\
40$\times$48      & \cmark  & \xmark       & .756$\pm$.039  & .753$\pm$.407 & 888\\
40$\times$48      & \cmark  & \cmark       & .756$\pm$.037  & $<$0.0001         & 888\\\hline
U-Net & --  & \cmark       & .762$\pm$.039  & $<$0.0001        & 2190\\\hline
\end{tabular}
\end{small}
% }
\caption{Ablation and parameter studies. SS denotes squaring and scaling, $|J|_{< 0}\%$ is the percentage of negative values of Jacobian determinant of deformation. MA(M) refers to the number of mult-adds operations in millions.}
\label{tab:ablation}   
\end{table}

We also prove the necessity of embedding a DFT layer in the encoder. Without this layer, our encoder is purely a CNN that has to learn complex-valued Fourier coefficients from image pairs. Following DeepFlash \cite{wang2020deepflash}, we use two networks to separately compute the real and imaginary parts of these complex coefficients. As reported in Table \ref{tab:ablation}, using the DFT layer, Dice is improved by 6.8\% and 8.1\% for the patch sizes of 20$\times$24 and 40$\times$48, respectively, which validates the efficacy of such a layer. This experiment shows the superiority of our proposed band-limited representation over DeepFlash's.

We further study the impact of adding a squaring and scaling module into Fourier-Net. As shown in Table \ref{tab:ablation}, this module encourages diffeomorphisms for the estimated deformation, due to the fact that it produces less percentage of negative values of Jacobian determinant of deformation.

\begin{table}[t]
\centering
% \resizebox{\columnwidth}{!}{
\begin{small}
\begin{tabular}{lcccc}\hline
Methods     & Patch & Dice$\uparrow$        & $|J|_{ < 0}\%$        & CPU \\\hline
Initial       & - & .544$\pm$.089 &-           &-     \\
Flash       & 16$\times$16      & .702$\pm$.051 & .033$\pm$.126  & 13.7  \\
Flash       & 20$\times$24      & .727$\pm$.046 & .205$\pm$.279 &  22.6     \\
Flash       & 40$\times$48      & .734$\pm$.045 & .049$\pm$.080 & 85.8     \\
DeepFlash   & 16$\times$16      & .615$\pm$.055 & 0.0$\pm$0.0            & .487         \\
DeepFlash   & 20$\times$24      & .597$\pm$.066            & 0.0$\pm$0.0 &.617         \\\hline
% DeepFlash   & 40x48      &             &             &     &     \\
B-Spline-Diff   & 20$\times$24      & .710$\pm$.041 &  .014$\pm$.072  &.012          \\
B-Spline-Diff   & 40$\times$48      & .737$\pm$.038  & .015$\pm$.039   &.012      \\\hline
F-Net$_{\textbf{S}}$ &40$\times$48     & .748$\pm$.039  & .671$\pm$.390             & \textbf{.007}      \\
F-Net-Diff$_{\textbf{S}}$ &40$\times$48     & .750$\pm$.038  & $<$0.0001             & .010      \\  
F-Net &40$\times$48     & .756$\pm$.039  & .753$\pm$.407             & .011         \\
F-Net-Diff &  40$\times$48 & .756$\pm$.037  & $<$0.0001   & .015        \\ 
F-Net${_{\textbf{L}}}$ &40$\times$48     & .759$\pm$.040  & .781$\pm$.406             & .037         \\
F-Net-Diff${_{\textbf{L}}}$ &  40$\times$48 & \textbf{.761$\pm$.037}  & 0.0$\pm$0.0   & .040        \\\hline
\end{tabular}
\end{small}
% }

\caption{Comparing different methods on 2D OASIS. F-Net is the abbreviation for Fourier-Net. All reported CPU runtimes (in seconds) are tested on the same machine.}

\label{tab:oasis2d}
\end{table}

\begin{table}[t]

\centering
% \begin{small}
    
% \resizebox{0.9\columnwidth}{!}{
\begin{tabular}{lccc}\hline
Methods      & Dice$\uparrow$         &HD95$\downarrow$  \\\hline
Initial       &.572$\pm$.053 & 3.831     \\
nnU-Net \cite{hering2022learn2reg}        & .846$\pm$.016  & 1.500     \\
LapIRN        & .861$\pm$.015  & 1.514   \\
TransMorph     &.858$\pm$.014 & 1.494   \\
TransMorph-Large     &\textbf{.862$\pm$.014}  &	1.431   \\ \hline
Fourier-Net-Diff &.843$\pm$.013  & 1.495                             \\
Fourier-Net &.847$\pm$.013  & 1.455 \\
Fourier-Net$_{\textbf{L}}$ & .860$\pm$.013  & \textbf{1.375}\\\hline
\end{tabular}
% }
% \end{small}
\caption{Performance comparison on 3D OASIS which is the MICCAI Learn2reg 2021 Task 3 validation dataset. All results  are taken from the leaderboard. HD95 is the 95\% Hausdorff distance, a lower value suggests a better performance.}
\label{tab:oasis3d}
\end{table}

\subsection{Comparison on Inter-subject Registration}
\textbf{2D OASIS:} In Table \ref{tab:oasis2d}, we compare the performance of Fourier-Net with Flash \cite{zhang2019fast}, DeepFlash \cite{wang2020deepflash}, and B-Spline-Diff \cite{qiu2021learning}. We manage to compile and run Flash\footnote{\url{https://bitbucket.org/FlashC/flashc/src/master/}} in CPU, but its official GPU version keeps throwing \textit{segmentation fault} errors. We report the performance of Flash on three band-limited patch sizes, and its built-in hyper-parameters are grid-searched over 252 different combinations on the whole validation set for each size. We also manage to run DeepFlash\footnote{\url{https://github.com/jw4hv/deepflash}} with supervision from Flash's results. We train DeepFlash on all 40200 training pairs for 1000 epochs with more than 40 different combinations of hyper-parameters and report the best results. B-Spline-Diff is also trained with all training pairs using its official implementation\footnote{https://github.com/qiuhuaqi/midir}.

 % provided by the authors i
%%% We managed to XXX (). The details for reproducing these three methods are provided in the supplementary material.

In Table \ref{tab:oasis2d}, all Fourier-Net variants outperform competing methods in terms of Dice. Specifically, our Fourier-Net$_{\textbf{S}}$ achieves a 0.748 Dice score with 0.007 seconds inference speed per image pair. Compared to Flash using a 40$\times$48 patch, Fourier-Net$_{\textbf{S}}$ improves Dice by 1.5\% and is 12,257 times faster. Though DeepFlash is much faster than Flash, we find that DeepFlash is very difficult to converge and as such achieves the lowest Dice score (0.597). Moreover, DeepFlash is not an end-to-end method, because its output (band-limited velocity field) requires an additional PDE algorithm to compute the final deformation. As such, it is much slower than deep learning methods such as ours or B-Spline-Diff (0.012 seconds per image pair on CPU). Note that the computational time is averaged on the whole test set, including the cost of loading models and images.

Note that the speed advantage of Fourier-Net on CPU decreases when we use larger models such as Fourier-Net$_{\textbf{L}}$, but its performance can be boosted by 1.1\% compared to Fourier-Net$_{\textbf{S}}$ in terms of Dice.

We also list the percentage of negative values of Jacobian determinant of deformation for all compared methods in Table \ref{tab:oasis2d}. Though both Flash and B-Spline-Diff are diffeomorphic approaches, neither of them produces perfect diffeomorphic deformations on this dataset. The proposed three Fourier-Net-Diff variants, however, barely generate negative Jacobian determinants and are therefore diffeomorphic.
 % For DeepFlash, we make slight modifications to adapt different input and output resolutions. 

% \begin{table}[]
% \caption{Comparisons on the MICCAI L2R 2021 Task3 Validation Dataset}
% \centering
% \label{tab:oasis3d}
% \resizebox{.50\textwidth}{!}{
% \begin{tabular}{cccc}\hline
% Methods      & Dice        &SDlogJ &HdDist95  \\\hline
% nnU-Net        & 0.8464$\pm$0.0159 & 0.0668 & 1.5003     \\
% LapIRN       & 0.8610$\pm$0.0148 & 0.0721 & 1.5139   \\
% TransMorph-Large     &0.8623$\pm$0.0144 &	0.1276 &	1.4315   \\
% TransMorph-Large-Cascade     &0.8691$\pm$0.0145& 0.0945 & 1.3969   \\\hline
% Fourier-Net-Diff &0.8429$\pm$0.0134 & 0.1117 & 1.4952                            \\
% Fourier-Net &0.8468$\pm$0.0134 & 0.2682 & 1.4547\\
% Fourier-Net$_{\textbf{L}}$ & 0.8604$\pm$0.0134 & 0.4782 & 1.3748\\\hline
% \end{tabular}}
% \end{table}

% The results suggest the superiority of the proposed Fourier-Net over these compared methods on both inference time and accuracy.

% Two different interpolation based methods are included, first, we compared the simple direct bi-linear interpolation to enlarge the displacement or velocity field to the full resolution, additionally, we included the well-known B-Spline interpolation.

\textbf{3D OASIS:} In Table \ref{tab:oasis3d}, we further compare Fourier-Net with other methods on the MICCAI 2021 Learn2reg challenge dataset. Though Fourier-Net is slightly lower than LapRIN\cite{mok2020large} in Dice, it achieves a better Hausdorff distance than LapRIN with a 0.059 improvement. If we use a larger Fourier-Net$_{\textbf{L}}$, it can achieve the lowest HD95, suggesting that Fourier-Net is able to obtain comparable results on par with state-of-the-art on this dataset.

\begin{table*}[ht]

\centering
% \resizebox{0.80\textwidth}{!}{
\begin{small}
\begin{tabular}{lcccccc}
\hline
Methods            & Dice$\uparrow$ & $|J|_{ < 0}\%$     & Parameters & Mult-Adds (G) & CPU (s) & GPU (s)\\\hline
Affine$^\ast$    & .386$\pm$.195 &- & -                      &-&-&-\\
SyN$^\ast$ \cite{avants_ANTS}       & .645$\pm$.152 & \textless{}0.0001      &-&-&-&-\\
NiftyReg$^\ast$ \cite{modat2010fast} & .645$\pm$.167 & \textless{}0.0001      &-&-&-&-\\
LDDMM$^\ast$ \cite{beg2005computing}     & .680$\pm$.135 & \textless{}0.0001      &-&-&-&-\\
Flash \cite{zhang2019fast}           & .692$\pm$.140 & 0.0$\pm$0.0            &-&-&- & -\\
deedsBCV$^\ast$   & .733$\pm$.126 & 0.147$\pm$0.050        &-&-&-&-\\\hline
VoxelMorph-1$^\ast$ \cite{balakrishnan2019voxelmorph}  & .728$\pm$.129 & 1.590$\pm$0.339    &274,387&304.05&9.373&0.391\\
VoxelMorph-2$^\ast$ \cite{balakrishnan2019voxelmorph}  & .732$\pm$.123 & 1.522$\pm$0.336    &301,411&398.81&10.530&0.441\\
VoxelMorph-Diff$^\ast$   & .580$\pm$.165 & \textless{}0.0001      &307,878&89.67 &3.691 & 0.418\\
B-Spline-Diff$^\ast$ \cite{qiu2021learning}    & .742$\pm$.128 & \textless{}0.0001      &266,387&47.05&7.076 & 0.437\\%\hline
TransMorph$^\ast$ \cite{chen2021transmorph}       & .754$\pm$.124 & 1.579$\pm$0.328        &46,771,251&657.64 & 22.035 & 0.443\\
% TransMorph-Bayes & 0.746$\pm$0.123 & & 1.560$\pm$0.333        &21,205,491&657.69\\
TransMorph-Diff$^\ast$ \cite{chen2021transmorph}  & .594$\pm$.163 & \textless{}0.0001      &46,557,414&252.61&10.389  &0.438\\
TransMorph-B-Spline$^\ast$ \cite{chen2021transmorph} & .761$\pm$.122 & \textless{}0.0001      &46,806,307&425.95 & 18.138 & 0.442\\\hline
Fourier-Net$_{\textbf{S}}$      & .759$\pm$.132 & 0.009$\pm$0.008       &1,050,800&  43.82 &\textbf{1.919} &\textbf{0.318}\\
Fourier-Net-Diff$_{\textbf{S}}$ & .756$\pm$.130 & 0.0$\pm$0.0      &1,050,800&  43.82 &6.202 &0.332\\
Fourier-Net      & \textbf{.763$\pm$.129} & 0.024$\pm$0.019       &4,198,352&  169.07 & 4.423 &0.342\\
Fourier-Net-Diff & .761$\pm$.131 & 0.0$\pm$0.0     &4,198,352&  169.07 &8.679 &0.345\\\hline
\end{tabular}
    
\end{small}
% }

\caption{Performance comparison between different methods on IXI.  Results of the methods labeled with $\ast$ are taken from TransMorph \cite{chen2021transmorph}, as we used the exact same data splitting and testing protocol as TransMorph. The reported runtimes of all deep methods are computed by us on the same machine and are averaged on the whole testing set.}
\label{tab:ixi}
\end{table*}

\begin{figure*}[!ht]
  \centering
  % include second image
  \includegraphics[width=0.98\linewidth]{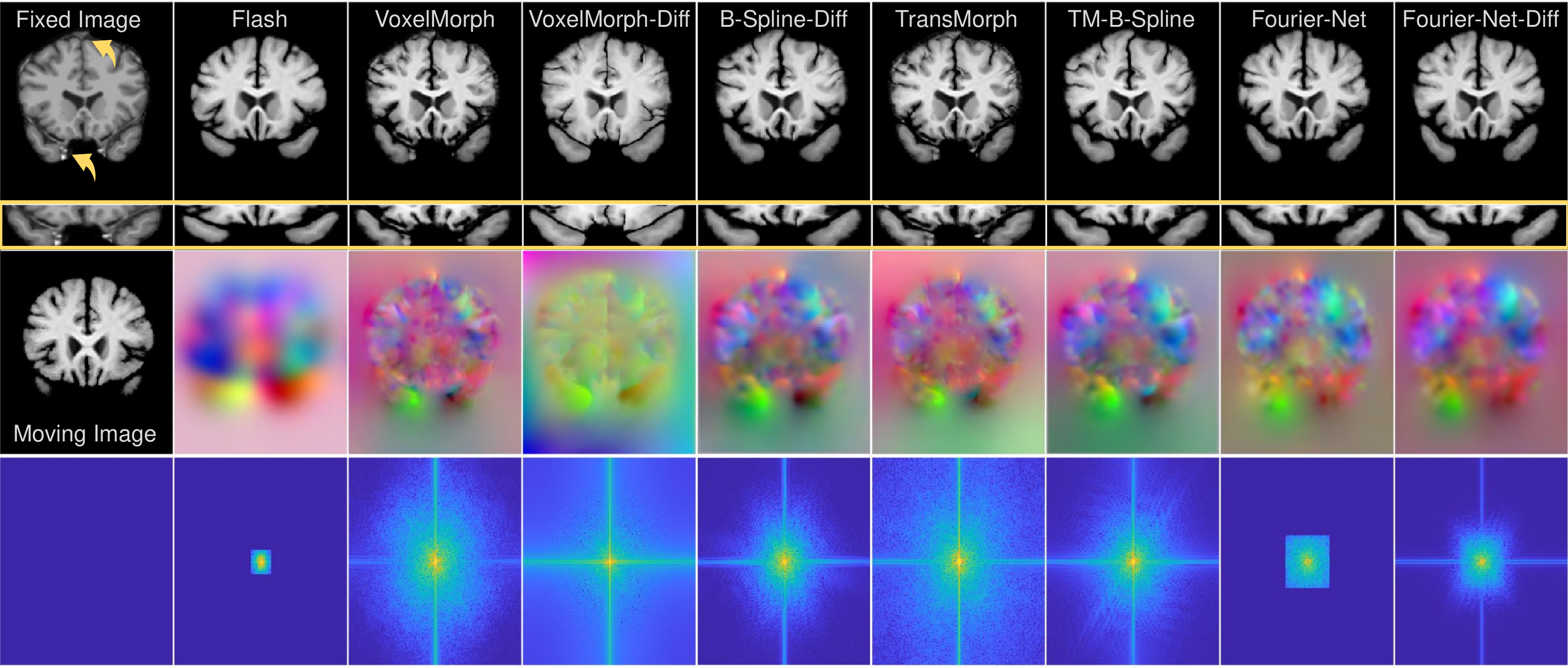}
  \caption{Visual comparison between different methods on 3D IXI. The 1st column displays a fixed image, a moving image, and a placeholder. From top to bottom rows excluding the 1st column: warped moving images (with a zoomed area in the yellow box), displacement fields, and displacement fields after DFT. Our Fourier-Net and Fourier-Net-Diff produce smoother deformations and better warped moving images (see noisy regions marked by yellow arrows and box).}
% \end{subfigure}
\label{fig:visualresult}
\end{figure*}

\subsection{Comparison on Atlas-Based Registration}
% , B-Spline-Diff \cite{qiu2021learning}
\textbf{3D IXI:} In Table \ref{tab:ixi}, we first compare our Fourier-Net with iterative methods such as Flash and deedsBCV \cite{heinrich2015multi} and deep learning methods such as TransMorph-B-Spline \cite{chen2021transmorph}, which is a combination of TransMorph and B-Spline-Diff. Note that for Flash, 200 combinations of hyper-parameters are grid-searched using 5 randomly selected validation samples. We do not include all images in the validation set because tuning Flash on CPU can take up to \textit{30 minutes} for each pair.

% Except Flash, the performance of rest iterative methods is taken from TransMorph \cite{chen2021transmorph}. 
% , ANTS SyN, NiftyReg \cite{modat2010fast}, LDDMM , and deedsBCV .
% As listed in Table \ref{tab:ixi},

The proposed Fourier-Net achieves the highest Dice score (0.763) with 4.42s inference speed per image pair. By using less number of kernels in each layer, Fourier-Net$_{\textbf{S}}$ achieves the fastest inference speed (1.92s) on CPU, which is faster than all other deep learning methods, while retaining a competitive accuracy. Furthermore, Fourier-Net$_{\textbf{S}}$ outperforms TransMorph by 0.5\% in Dice with only 2.2$\%$ of its parameters and 6.66$\%$ of its mult-adds. In terms of inference speed, Fourier-Net$_{\textbf{S}}$ is 11.48 times faster than TransMorph (22.035 seconds) on CPU. Finally, Table \ref{tab:ixi} (3$^{\text{rd}}$ column) indicates that Fourier-Net-Diff barely generates any folding and thus effectively preserves diffeomorphisms.

% Moreover, our Fourier-Net is also the fastest method among all compared methods when tested on GPU. The inference speed of Fourier-Net-Diff is slightly slower than B-Spline-Diff on CPU, but Fourier-Net only takes 4.42s per image pair and is 2.65s faster than B-Spline-Diff. 

 From Figure~\ref{fig:visualresult},  we can observe that Flash's deformation also has no foldings, but it over-smoothes its displacement field, resulting in a less accurate warping. Figure~\ref{fig:visualresult} (last row) shows that only Flash and Fourier-Net produce strictly band-limited Fourier coefficients, and that the deformation of Fourier-Net-Diff is no longer band-limited due to the use of the squaring and scaling layers.
\section{Conclusion}
In this paper, we propose to learn a low-dimensional representation of displacement/velocity field in the band-limited Fourier domain. Experimental results on two brain datasets show that our Fourier-Net is more efficient than state-of-the-art methods in terms of speed while retaining a comparative performance in terms of registration accuracy.

\newpage
\section*{Acknowledgements}
\noindent The computations described in this research were performed using the Baskerville Tier 2 HPC service. Baskerville was funded by the EPSRC and UKRI (EP/T022221/1 and EP/W032244/1) and is operated by Advanced Research Computing at the University of Birmingham. Xi Jia is partially supported by the Chinese Scholarship Council.

\bibliography{aaai23}
\end{document}